\documentclass{article}

\usepackage[preprint]{neurips_2025}

\usepackage{enumitem}
\usepackage[utf8]{inputenc} 
\usepackage[T1]{fontenc}    
\usepackage[colorlinks,linkcolor={blue!60!black},citecolor={blue!60!black},urlcolor={blue!60!black}]{hyperref}       
\usepackage{url}            
\usepackage{booktabs}       
\usepackage{amsfonts}       
\usepackage{nicefrac}       
\usepackage{microtype}      
\usepackage[dvipsnames]{xcolor}         
\usepackage{cancel}
\usepackage{algorithm}
\usepackage{algpseudocode}
\usepackage{booktabs,siunitx}
\usepackage{graphicx}
\usepackage{csquotes}

\usepackage{amsmath,amsfonts,bm}

\def\Wh{{\bm{\hat{W}}}} 
\newcommand{\Whi}[1]{{\hat{W}_{#1}}} 
\newcommand{\Wp}{{\mW'}}
\newcommand{\Wpi}[1]{W_{#1}'}

\newcommand{\Hp}{{\mH'}}
\newcommand{\Hip}{{\left(\mH'\right)^{-1}}}
\newcommand{\mCpi}[1]{{\mC}_{#1}'}
\newcommand{\Cpi}[1]{{C}_{#1}'}
\newcommand{\Cp}{\mC'}

\newcommand{\Hipi}[2]{{[(\mH'_{\geq #1,\geq #1})^{-1}]}_{#2}}

\def\eqref#1{equation~\ref{#1}}

\def\1{\bm{1}}

\def\mA{{\bm{A}}}

\def\mC{{\bm{C}}}

\def\mH{{\bm{H}}}
\def\mI{{\bm{I}}}

\def\mW{{\bm{W}}}
\def\mX{{\bm{X}}}

\DeclareMathAlphabet{\mathsfit}{\encodingdefault}{\sfdefault}{m}{sl}
\SetMathAlphabet{\mathsfit}{bold}{\encodingdefault}{\sfdefault}{bx}{n}

\DeclareMathOperator{\Tr}{Tr}

\usepackage{hyperref}
\usepackage{url}
\usepackage{caption}
\usepackage{multirow}
\usepackage{wrapfig}

\usepackage{placeins}

\DeclareMathSymbol{:}{\mathord}{operators}{"3A}

\DeclareMathOperator*{\argmin}{\arg\min}

\algrenewcommand\algorithmicrequire{\textbf{Input:}} 
\algrenewcommand\algorithmicensure{\textbf{Output:}} 
\newcommand{\Mycomment}[1]{{\color{darkgray}\hypersetup{linkcolor=blue!30!darkgray}\Comment{\emph{#1}}}} 
\algrenewcommand\algorithmicindent{0.9em} 

\usepackage{hyperref}
\newcommand{\algoref}[1]{\hyperref[#1]{Algorithm~\ref*{#1}}}

\title{Reducing Storage of Pretrained Neural Networks by Rate-Constrained Quantization and Entropy Coding}

\author{%
  Alexander Conzelmann \\
  Department of Computer Science\\
  University of Tübingen\\
  Tübingen, Germany\\
  \texttt{a.conzelmann@uni-tuebingen.de} \\
  \And
  Robert Bamler \\
  Department of Computer Science\\
  University of Tübingen\\
  Tübingen, Germany\\
  \texttt{robert.bamler@uni-tuebingen.de} 
}

\begin{document}

\maketitle

\begin{abstract}
The ever-growing size of neural networks poses serious challenges on resource-constrained devices, such as embedded sensors. Compression algorithms that reduce their size can mitigate these problems, provided that model performance stays close to the original.
We propose a novel post-training compression framework that combines rate-aware quantization with entropy coding by (1) extending the well-known layer-wise loss by a quadratic rate estimation, and (2) providing locally 
exact solutions to this modified objective following the Optimal Brain Surgeon (OBS) method. 
Our method allows for very fast decoding and is compatible with arbitrary quantization grids. 
We verify our results empirically by testing on various computer-vision networks, achieving a 20-40\% decrease in bit rate at the same performance as
the popular compression algorithm NNCodec.

Our code is available at \url{https://github.com/Conzel/cerwu}.
\end{abstract}

\section{Introduction}
While neural networks have achieved impressive results on various tasks, their large computational demands require many neural network applications to  
rely on server-side deployment. This increases latency and can cause concerns for end users regarding privacy and regulatory constraints \citep{saravananAdvancementsOnDeviceDeep2023,hohmanModelCompressionPractice2024}. Thus, there has been an increased effort to move 
machine learning pipelines back onto user devices, which usually involves techniques that reduce their resource demands \cite{hohmanModelCompressionPractice2024}. The works on neural network compression can be roughly divided into two groups: 
(1) works that aim to reduce the \textit{server-side} costs of training \cite{liMemoryEfficientOptimizers2023,haoNeuZipMemoryEfficientTraining2024}, 
fine-tuning \citep{dettmersQLoRAEfficientFinetuning2023a,qinAccurateLoRAFinetuningQuantization2024} or inference \citep{kwonEfficientMemoryManagement2023,hooperKVQuant10Million2024,parkInferenceOptimizationFoundation2024}. These methods often save computational costs or reduce GPU memory requirements.
And (2) works that focus on \textit{embedded and edge devices}, 
aiming to produce extremely small models still suitable for inference \citep{howardSearchingMobileNetV32019,guMiniLLMKnowledgeDistillation2023}, to reduce
energy demands \citep{kroukaEnergyEfficientModelCompression2021,baskinCATCompressionAwareTraining2021} or to decrease the storage requirements \citep{choiUniversalDeepNeural2020,beckingNNCodecOpenSource2023}.
While both groups use similar techniques, such as pruning, quantization and entropy coding, the methods might not be used interchangeably, as they target different architectures and scenarios.

Our work explicitly focuses on the storage requirements of neural networks, which we argue is important for more widespread adoption of neural networks on edge devices.
These devices often have limited storage capacity or network bandwidth for software updates,
causing model-size issues with even the simplest of neural networks (even ResNets~\citep{heDeepResidualLearning2016} have an uncompressed storage size of hundreds of megabytes).
The importance of storage size has been reinforced with 
the introduction of the ISO/IEC 15938-17 standard, a technical standard for the compression of neural networks.

So far, few other works explicitly target the 
storage size of neural networks.
When focusing just on inference speed, quantization 
(i.e., saving each parameter with a lower precision) is often enough, as current GPU kernels only experience a speed-up for 4-bit or 8-bit quantization (see \citep{nagelWhitePaperNeural2021}).
However, to reduce storage size even further, a more powerful technique is to use entropy coding, which builds a 
probabilistic model of the data and then encodes the data points
into bit strings whose length is proportional to their \emph{information content} (negative log-probability), 
allowing us to encode parameters with \enquote{fractional bits}, 
in some cases even achieving rates below 1~bit per weight.

In this work, we combine quantization with entropy coding,
and propose a novel approach that produces highly
compressible quantized layer representations by adding a quadratic 
rate-estimation to the layer-wise loss~\citep{nagelAdaptiveRoundingPostTraining2020}, searching
for rate-distortion optimal quantization choices, and then providing
entropy-regularized optimal weight updates following the Optimal Brain Surgeon (OBS)~\citep{hassibiOptimalBrainSurgeon1993} framework.
In summary, our proposed network compression scheme
\begin{itemize}[itemsep=-2pt, topsep=-3pt]
    \item achieves extremely low storage cost for the compressed networks;
    \item allows for extremely fast decoding of compressed networks;
    \item produces quantized representations
    suitable for fast inference using integer arithmetic; and
    \item is flexible in regards to the choice of quantization grid and entropy model used, allowing users to, e.g., trade off decompression or inference speed against model performance.
\end{itemize}

We name our method \textbf{CERWU} (\textbf{C}ompression with \textbf{E}ntropy-\textbf{R}egularized \textbf{W}eight \textbf{U}pdates). In the rest of this paper, we discuss related work (\autoref{sec:related}) and the information-theoretical background (\autoref{sec:background}), 
derive the proposed entropy-regularized weight updates (\autoref{sec:method}), and verify the effectiveness of our method empirically (\autoref{sec:experiments}) on various networks from the computer vision community.
We conclude with a discussion of limitations of our method (\autoref{sec:conclusion}).

\section{Related Work}
\label{sec:related}
Various target scenarios exist in network compression.
Our method is a \textit{post-training} compression method that 
reduces the \textit{storage size} of a network. Other compression schemes focus instead on decreasing the memory footprint on GPUs 
\cite{shengFlexGenHighThroughputGenerative2023,kwonEfficientMemoryManagement2023}, 
reducing energy costs \citep{koAdaptiveWeightCompression2017,chmielFeatureMapTransform2019,kroukaEnergyEfficientModelCompression2021}, 
increasing inference speed \cite{liPENNIPrunedKernel2020,dettmersLLMInt88bit2022} or they require re-training for each tested compression setting~\cite{baskinCATCompressionAwareTraining2021,chenEfficientQATEfficientQuantizationAware2024}.

Most related work focuses on network \emph{quantization} rather than compression (often aiming to improve inference speed).
We discuss the difference between the two in \autoref{sec:compress-vs-quantize} and existing quantization methods in \autoref{sec:quantization}.
There are fewer existing works with our focus on storage size.
Han et al. \citep{hanDeepCompressionCompressing2016} combine pruning, quantization, weight-sharing and entropy coding to achieve very high compression factors. 
Universal Neural Network Compression \citep{choiUniversalDeepNeural2020} uses universal quantization \citep{zivUniversalQuantization1985} together with vector quantization. Wiedemann et al. \cite{wiedemannCompactComputationallyEfficient2020} propose to use an entropy-constraint together with a sparsification process, resulting in parameter representations that allow for high compression ratios.

The \textit{Neural Network Compression and Representation Standard} 
(ISO/IEC 15938-17)~\citep{kirchhofferOverviewNeuralNetwork2022}, with its reference implementation NNCodec~\citep{beckingNNCodecOpenSource2023}, defines a compression pipeline encompassing parameter reduction (pruning, sparsification, parameter sharing, et cetera), quantization, and entropy coding, as well as interoperability with well-known neural network exchange formats such as ONNX \citep{bai2019}. 

\section{Background}
\label{sec:background}

\subsection{Compression vs.\ Quantization}
\label{sec:compress-vs-quantize}

Neural network compression is often conflated with quantization, but the two are fundamentally different.
In its most basic form, quantization reduces the precision of each network weight to a fixed (integer) bit-width (e.g., 4~or 8 bits per weight).
By contrast, compression is a more global operation that aims to find the most compact binary representation of a network by packing the most relevant information contained in all weights into a bit string of shortest possible length (called \emph{bit~rate}).
Compression thus effectively maps weights to varying bit-widths, and modern compression methods \citep{rissanen1979arithmetic, pasco1976source, martin1979range, duda2015use, bamler2022understanding} are not even constrained to bit boundaries, i.e., they can effectively assign fractional bit-widths to weights, including widths below 1~bit per weight, see \autoref{sec:inf-theory} below.

The simplest way to combine quantization and compression is to first quantize the weights to a fixed bit-width and then apply lossless compression.
However, this is empirically far from optimal (see ablation ``CERWU-$\lambda\!=\!0$'' and baseline ``RTN+EC'' in \autoref{sec:experiments}) because quantizing without consideration of the compression mechanism produces poorly compressible quantized weights.

Recent literature includes more advanced network quantization schemes that go beyond a fixed bit-width per weight, either by combining quantization with pruning \citep{frantarCompressionScalingLaws2025}, or by reserving extra bits for a small percentage of highly salient or outlier weights~\cite{dettmersLLMInt88bit2022,dettmersSpQRSparseQuantizedRepresentation2023}.
However, both the tuning of these extra steps and the binary representation of the required metadata (i.e., how the matrix indices of the pruned or salient weights can be compactly stored in a file) are usually ad-hoc and suboptimal as explicit minimization of the resulting bit rate is infeasible.
By contrast, an information-theoretically grounded approach, as proposed in this paper, subsumes both pruning and saliency-dependent quantization as limiting cases, and allows us to interpolate between them by explicitly minimizing a trade-off between the introduced error and the resulting overall bit rate after compression.
The next subsection introduces the relevant concepts from information theory and compression that we use in our work.

\subsection{Lossy Compression and Information Theory}\label{sec:inf-theory}

We consider, for simplicity, the problem of compressing the weight matrix~$\mW \in \mathcal W$ of a single layer of a neural network, where $\mathcal W = \mathbb R^{n \times m}$ for some input and output dimensions $m$ and~$n$, respectively.
The goal of lossy compression is to find an encoder~$e$ that maps~$\mW$ to a short bit string $e(\mW) \in {{\{0,1\}}^*} \,{:\!=}\, \bigcup_{\ell=0}^\infty {\{0,1\}}^\ell$, and a corresponding decoder~$d$ that maps this bit string to a reconstruction $\Wh \,{:\!=}\, d(e(\mW)) \in \hat{\mathcal W}$, such that $\Wh$ resembles $\mW$ with only a small error.%
\footnote{More general formulations admit for stochastic en-/decoders. But without an additional constraint such as realism, there always exists a pair of deterministic en-/decoders among the minimizers of \autoref{eq:rate-distortion-orig}.}
Here, $\hat{\mathcal W} \subset \mathcal W$ is a typically discrete reconstruction space (discussed in \autoref{sec:quantization} below).
To make this problem well-defined, one has to specify a distortion metric $D:\, {\mathcal W \times \hat{\mathcal W} \to \mathbb R_{\geq0}}$ that quantifies the reconstruction error, and a probabilistic model~$P_\mW$ of the data source, which models prior assumptions that the decoder is allowed to make about the weights before looking at the compressed bit string (for example, the prior assumption that $\mW$ is approximately sparse could be expressed by a~$P_\mW$ that models the weights as i.i.d.\ with a sharp peak around zero).
Introducing a Lagrange parameter $\lambda\geq 0$ that trades off between bit rate and distortion, the optimal encoder/decoder pair $(e_*, d_*)$ minimizes the following \textit{rate-distortion objective} (where $|\cdot|$ is the length of the bit string):
\begin{align}\label{eq:rate-distortion-orig}
    \operatorname{RD}(\lambda) {\,:\!=\,} \min_{e,d}  \mathbb E_{\mW\sim P_\mW}\Bigl[
        D\bigl(\mW,d(e(\mW))\bigr)
        +\lambda \bigl|e(\mW)\bigr|\Bigr].
\end{align}
Here, the objective takes an expectation over $P_\mW$ even though we are ultimately only interested in compressing a specific weight matrix~$\mW$.
This ensures that the resulting decoder~$d_*$ that minimizes \autoref{eq:rate-distortion-orig} only uses the prior assumptions expressed by~$P_\mW$, preventing it from simply storing~$\mW$.

\paragraph{Information content and entropy coding.}
The minimization problem in \autoref{eq:rate-distortion-orig} is intractable in practice, but we can simplify it by splitting it into a \emph{quantization} step and an \emph{entropy coding} step, where the latter can be solved efficiently~\citep{mackay2003information}.
Observe that an optimal decoder~$d_*:\, {{{\{0,1\}}^*} \to \hat{\mathcal W}}$ that solves the minimization problem in \autoref{eq:rate-distortion-orig} is invertible since an encoder/decoder pair that reserves two different bit strings for the same reconstruction~$\Wh$ could reduce its expected bit rate by reassigning one of the two bit strings to some $\Wh'\neq \Wh$ that is currently assigned to a longer bit string.
Thus, with the substitution ${q(\mW) {\,:\!=\,} d(e(\mW))}$, \autoref{eq:rate-distortion-orig} simplifies to
\begin{align}\label{eq:rate-distortion-orig-split}
    \operatorname{RD}(\lambda)
    &= \min_{q,d}  \mathbb E_{\mW\sim P_\mW}\Bigl[
        D\bigl(\mW,q(\mW)\bigr)
        +\lambda \bigl|d^{-1}(q(\mW))\bigr|\Bigr]
\end{align}
where $d^{-1}(q(\mW)) = e(\mW)$ splits the encoder into a \emph{quantizer} $q:\, {\mathcal W \to \hat{\mathcal W}}$ and an \emph{entropy coder} $d^{-1}:\, {\hat{\mathcal W} \to {\{0,1\}}^*}$, which now no longer appers in the distortion term.
Here, the literature on lossless compression provides two important insights.
First, the source coding theorem \citep{shannonMathematicalTheoryCommunication1948,mackay2003information} states that the bit rates of an \emph{optimal} entropy coder $d_*^{-1}$ that minimizes $\mathbb E_{P_\mW}[|d^{-1}(q(\mW))|]$ in \autoref{eq:rate-distortion-orig-split} for a given quantizer~$q$ are given (up to at most 1~bit) by the \emph{information content} of the quantized weights,
\begin{align}\label{eq:source-coding-theorem}
    |d_*^{-1}(\Wh)| = -\log_2 P_{\Wh}(\Wh) + \epsilon
    \qquad\forall\, \Wh \in \hat{\mathcal W}
\end{align}
where $\epsilon <1$ is negligible for any realistically large total bit rate, and the \emph{entropy model}~$P_{\Wh}$ is the push-forward of~$P_\mW$ along~$q$ (in practice, our proposed method directly models $P_\Wh$ instead of~$P_\mW$).

The second important insight from the lossless compression literature is that the theoretically optimal bit rate in \autoref{eq:source-coding-theorem} can very nearly be achieved in practice by computationally efficient entropy coding algorithms such as arithmetic coding~\citep{rissanen1979arithmetic, pasco1976source}, range coding~\citep{martin1979range}, or ANS \citep{duda2015use, bamler2022understanding}.
Thus, the minimization over~$d$ in \autoref{eq:rate-distortion-orig-split} is solved, and the remaining task is to find a quantizer~$q$ that minimizes $\mathbb E_{P_\mW}\bigl[ D(\mW,q(\mW)) - \lambda \log_2 P_{\hat\mW}(q(\mW)) \bigr]$.
Different to Equations~\ref{eq:rate-distortion-orig} and~\ref{eq:rate-distortion-orig-split}, this problem factorizes over~$\mW\in\mathcal W$, and so we only need to consider it for the specific weight matrix~$\mW$ that we actually want to compress.
Substituting $\Wh = q(\mW)$, we arrive at the rate/distortion objective
\begin{align}\label{eq:rate-distortion}
    \Wh_* = \argmin_\Wh \Bigl[ D(\mW,\Wh) + \lambda R(\Wh) \Bigr]
    \qquad\text{with the \emph{rate} $R(\Wh) = - \log_2 P_{\hat\mW}(\Wh)$.}
\end{align}

\paragraph{Fractional and sub-1-bit (amortized) bit rates.}
\autoref{eq:source-coding-theorem} provides an analytic expression for the bit rate of an optimal entropy coder.
It is hard to overstate the importance of this result as it allows us to attribute how much each matrix element~$\hat W_{ij}$ contributes to the total bit rate, even though an optimal entropy coder typically ``packs'' multiple matrix elements together.
For an autoregressive entropy model $P_{\hat\mW}(\hat\mW) = \prod_{i,j} P_{\hat\mW}(\hat W_{ij} \,|\, \hat W_{<(ij)})$ (where the notation ``${<\!(ij)}$'' assumes that some ordering is defined), the total bit rate for encoding the matrix~$\hat\mW$ splits into a sum, $|d_*^{-1}(\hat\mW)| = -\sum_{i,j} \log_2 P_{\hat\mW}(\hat W_{ij} \,|\, \hat W_{<(ij)}) + \epsilon$, where $\epsilon < 1$ appears only once outside the sum and can thus be neglected.
We can therefore interpret each term in this sum as the (amortized) contribution of an individual matrix element to the total bit rate, and minimizing these individual contributions minimizes the total bit rate~$|d_*^{-1}(\hat\mW)|$.
Importantly, these individual amortized bit rates are meaningful even though they are generally non-integer values and are often even below 1~bit in our experiments.
Even though such fractional bit rates could not be measured individually by explicitly constructing an optimal entropy coder and encoding a single quantized matrix element, minimizing them is still meaningful in the sense that, e.g., reducing 100 amortized bit rates by 0.3~bit each would reduce the total bit rate by 30~bit.
Being able to quantify bit rates with fractional resolution via \autoref{eq:source-coding-theorem} is crucial for obtaining good compression performance in the low-bit-rate regime.

\subsection{Quantization}
\label{sec:quantization}

While constructing a near-optimal entropy coder $d^{-1}$ in \autoref{eq:rate-distortion-orig-split} for a given entropy model is a solved problem, efficiently finding an optimal quantizer~$q$ is still unsolved.
Quantization maps the uncompressed weight matrix~${\mW\in\mathbb{R}^{n\times m}}$ to~$\Wh = q(\mW)$ in a discrete reconstruction space~$\hat{\mathcal W}$.

\paragraph{Quantization grid.}
Restricting~$\hat{\mathcal W}$ to a discrete (i.e., finite or countably infinite) set is unavoidable in (deterministic) compression because the decoder~$d$ maps to~$\hat{\mathcal W}$ from the space of bit strings~${\{0,1\}}^*$, which is countable.
To make our compression method compatible with inference acceleration methods that use integer arithmetic~\cite{nagelWhitePaperNeural2021}, our experiments use $\hat{\mathcal W} = G^{n \times m}$, with a symmetric, uniform grid
\begin{align}\label{eq:grid}
    G = \bigl\{ i \cdot {\|\mW\|}_{\infty} /( (k-1)/2) \bigr\}_{i=-(k-1)/2}^{(k-1)/2}
    \qquad\text{for some \emph{odd} grid size~$k$ (so that $0\in G$).}
\end{align}
Here, the grid size~$k=|G|$ controls how faithful quantization can be in the best case.
In the literature \citep{frantar2023optq,dettmersCase4bitPrecision2023,tsengQuIPEvenBetter2024}, $|G|=2^r$~is instead often restricted to be a power of two, in which case $r$ would be the number of bits that each quantized matrix element would occupy if its index into~$G$ was stored in uncompressed form.
This restriction to powers of two is not necessary in our compression method as we use entropy coding to reduce the storage cost of each weight to its actual information content.

More sophisticated designs~\cite{dettmersQLoRAEfficientFinetuning2023a,yang2020variational} place more grid points in high density regions of the data distribution, 
but this usually requires one to explicitly dequantize all quantized matrices before performing operations on them, preventing the use of accelerated inference operations. 

\paragraph{Quantization methods.}
The quantization problem in \autoref{eq:rate-distortion} is a high-dimensional discrete optimization problem, which is infeasible to solve exactly except for very simple distortion metrics~$D$ and entropy models~$P_{\hat\mW}$.
Various existing methods can be understood as approximations to this problem.
The crudest approximation is \emph{round to nearest} (RTN), i.e., independently setting $\hat W_{ij} = \argmin_{g\in G} (W_{ij} - g)^2$ for each $i$,~$j$.
Thus, RTN solves \autoref{eq:rate-distortion} for $\lambda=0$ and a distortion metric that factorizes over all elements of~$\Wh$.
NNCodec~\citep{beckingNNCodecOpenSource2023} extends the approach to $\lambda>0$ with the autoregressive entropy model DeepCABAC~\cite{wiedemannDeepCABACUniversalCompression2020}, constructing a greedy approximation to the optimal quantizer~$q$, i.e., when quantizing each matrix element~$W_{ij}$, NNCodec neglects the effect that the choice of~$\hat W_{ij}$ has on the bit rates of subsequent matrix elements by changing the internal state of the autoregressive entropy model.
Finally, OPTQ~\citep{frantar2023optq} implicitly considers $\lambda=0$ again, but uses as distortion function the non-factorized layerwise loss from~\citep{nagelAdaptiveRoundingPostTraining2020}, $D(\mW, \Wh) {\,:\!=\,} \| \mW \mX - \Wh \mX \|^2_2$, which considers the euclidean distance of layer outputs rather than of the weights themselves.
Here, $\mX$~is the layer input when evaluating the model on a so-called calibration set.
OPTQ approximates the minimum of $D(\mW, \Wh)$ over~$\Wh$ by iterating over the matrix elements in a fixed order and greedily rounding each element~$W_{ij}$ to the nearest neighbor in~$G$, but it then takes some aspects of the global structure of~$D$ into account by updating the remaining (so far unquantized) matrix elements of~$\mW$ to compensate, as well as possible, for the error introduced by quantizing~$W_{ij}$.

Our proposed method CERWU (Compression with Entropy-Regularized Weight Updates), presented in the next section, considers the full rate/distortion objective in \autoref{eq:rate-distortion} with $\lambda>0$ and a non-factorized distortion metric~$D$, and it takes the rate term into account both when quantizing each individual weight~$W_{ij}$, as well as (in an approximate way) during weight updates.

\section{Method}\label{sec:method}

Our proposed CERWU method starts from the rate/distortion-constrained quantization problem in \autoref{eq:rate-distortion}, where we use as the distortion $D$ the layer-wise loss popularized by~\citep{nagelAdaptiveRoundingPostTraining2020},
\begin{equation}\label{eq:layerwise-loss}
    \Wh_* = \argmin_\Wh L_\lambda(\Wh)
    \qquad\text{with}\qquad
    L_\lambda(\Wh) = \| \mW \mX - \Wh \mX \|^2_2 +  \lambda R(\Wh),
\end{equation}
where $\mX \in \mathbb R^{m \times p}$ is the layer input when evaluating the model on a calibration set of size~$p$, and the rate $R(\Wh) = -\log_2 P_\Wh(\Wh)$ assumes some entropy model~$P_\Wh$.
The rate term makes $L_\lambda(\Wh)$ non-quadratic, which complicates its minimization.
We therefore split $L_\lambda(\Wh) = L'_\lambda(\Wh) + L''_\lambda(\Wh)$ into a quadratic part $L'_\lambda(\Wh)$ that approximates $P_\Wh(\Wh)$ with (a discretization of) a Gaussian fit $\prod_{ij}\mathcal N\bigl(0, \operatorname{Var}(\{W_{ij}\}_{ij})\bigr)$ to the unquantized weights~$\{W_{ij}\}_{ij}$, and a remainder $L''_\lambda(\Wh)$,
\begin{align}\label{eq:regularized-ll}
    L'_\lambda(\Wh) = \| \mW \mX - \Wh \mX \|^2_2 + \frac{\lambda\gamma}{2} \|\Wh\|_2^2
    \qquad\text{and}\qquad
    L''_\lambda(\Wh) = \lambda R(\Wh) - \frac{\lambda\gamma}{2} \|\Wh\|_2^2,
\end{align}
with $\gamma = 1\big/\bigl(\ln(2) \operatorname{Var}(\{W_{ij}\}_{ij})\bigr)$.
We can now simplify $L_{\lambda}'(\Wh)$ by completing the square,%
\begin{align}\label{eq:regularized-ll-simple}
    L'_\lambda(\Wh) &= \frac12 \Tr\Bigl[ (\Wp - \Wh) \Hp (\Wp - \Wh)^T \Bigr] + \text{const.},
\end{align}
where the constant is independent of $\Wh$.
\autoref{eq:regularized-ll-simple} can be verified by multiplying out its r.h.s.\ (see Appendix~\ref{app:completing-the-square}) and comparing the result to $L'(\Wh)$ in \autoref{eq:regularized-ll}, using the entropy-regularized layer-wise Hessian~$\Hp$, the unregularized layer-wise Hessian~$\mH$, and the regularized weight matrix~$\Wp$,
\begin{align}\label{eq:renaming}
    \Hp = \mH + \lambda\gamma \mI, \qquad
    \mH = 2\mX\mX^T, \qquad\text{and}\qquad
    \Wp = \mW\mH(\Hp)^{-1}.
\end{align}

To obtain a quantized matrix $\Wh\in G^{n\times m}$ (the uniform grid~$G$ is defined in \autoref{eq:grid}) that approximately minimizes $L_\lambda(\Wh) = L'_\lambda(\Wh) + L''_\lambda(\Wh)$, we now iteratively apply the optimal brain surgeon (OBS) algorithm~\cite{hassibiOptimalBrainSurgeon1993} to the quadratic part $L'_\lambda(\Wh)$, interleaving it with a rate-constrained quantization step that takes the non-quadratic part $L''_\lambda(\Wh)$ into account as well.

\paragraph{Entropy-regularized weight updates.}
\algoref{algo:cerwu} summarizes our proposed CERWU algorithm.
Following~\cite{frantar2023optq}, we iterate over the rows~$i$ and columns~$j$ of the weight matrix without optimizing over the iteration order.
For each $(i,j)$, we first obtain $\Whi{ij} \in G$ by quantizing~$\Wpi{ij}$ as described below, and we then update so-far unquantized matrix elements of~$\Wp$ to optimally compensate for the error introduced by quantizing~$\Wpi{ij}$.
Since a closed-form solution for this update is only available for a quadratic loss function, we consider only the quadratic approximation $L'_\lambda(\Wh)$ in \autoref{eq:regularized-ll-simple} for the weight update, assuming that the non-quadratic remainder $L''_\lambda(\Wh)$ is small by construction.
Adapting the results in~\cite{hassibiOptimalBrainSurgeon1993} (see also derivation in \autoref{app:derivation}) to our quadratic loss $L'_\lambda(\Wh)$, only weights~$\mW'_{i,>j}$ in the same row~$i$ need to be updated, and the optimal update $\Delta_{\mW'_{i,>j}}$ and the incurred increase~$\Delta_{L'}$ of the quadratic part of the loss due to quantizing $\Wpi{ij}$ and updating $\mW'_{i,>j}$ are
\begin{equation}\label{eq:obs-update}
    \Delta_{\mW'_{i,>j}} = -\frac{\Wpi{ij} - \Whi{ij}}{\Hipi{j}{jj}}\, \Hipi{j}{j,>j}
    \qquad\text{and}\qquad
    \Delta_{L'} = \frac{1}{2} \frac{(\Wpi{ij} - \Whi{ij})^{2}}{\Hipi{j}{jj}}.
\end{equation}

Note that the weight update $\Delta_{\mW'_{i,>j}}$ (as well as the initialization of $\Wp$ in \autoref{eq:renaming}) depends on the \emph{entropy-regularized} Hessian~$\Hp$, which contains the term $\lambda\gamma\mI \propto \lambda/\operatorname{Var}(\{W_{ij}\}_{ij})$ that becomes large for a strong rate constraint~$\lambda$ and for small unquantized weights.
This regularization prevents the weights from accumulating large values over the course of repeated weight updates, which can hurt compressibility (see ablation ``CERWU-$\gamma\!=\!0$'' in \autoref{sec:experiments}).

Following \cite{frantar2023optq} (see also derivation in \autoref{app:proposition-iii}), we pre-compute all required inverse Hessian entries in \autoref{eq:obs-update} using the Cholesky decomposition $\Cp {\,:\!=\,} \text{Cholesky}(\Hip)^T$ to improve runtime and numerical stability, resulting in the equivalent equations
\begin{equation}\label{eq:optq-correction}
    \Delta_{\mW'_{i,>j}} = -\frac{\Wpi{ij} - \Whi{ij}}{\Cpi{jj}} \mCpi{j,>j}
    \qquad\text{and}\qquad
    \Delta_{L'} = \frac{1}{2} \frac{(\Wpi{ij} - \Whi{ij})^{2}}{\Cpi{jj}^2}.
\end{equation}

\paragraph{Quantization.}
For each $(i,j)$, before updating future weights~$\mW'_{i,>j}$, we quantize the current weight $\Wpi{ij}$ to $\Whi{ij} \in G$.
Since the grid size~$|G|$ is fixed, we can afford to explicitly search over all $g\in G$ and thus consider, in addition to $\Delta_{L'}$ from~\autoref{eq:optq-correction}, also changes to the non-quadratic part $L''(\Wh)$ of the layer-wise loss (\autoref{eq:regularized-ll}).
However, only the contribution of the current matrix entry $\Whi{ij}$ to $L''(\Wh)$ can be taken into account efficiently, and we have to neglect second-order effects to $L''(\Wh)$ due to resulting updates for~$\mW'_{i,>j}$.
Assuming an autoregressive entropy model $P_{\hat\mW}(\hat\mW) = \prod_{i,j} P_{\hat\mW}(\hat W_{ij} \,|\, \hat W_{<(ij)})$, the contribution of $\Whi{ij}$ to the rate $R(\Wh) = -\log_2 P_{\hat\mW}(\hat\mW)$ is $-\log_2 P_{\hat\mW}(\hat W_{ij} \,|\, \hat W_{<(ij)})$, and thus our quantization~$q$ sets $\Whi{ij} \gets q\bigl(\Wpi{ij}, \lambda, \gamma, P_{\hat\mW}(\;\cdot\; |\, \hat W_{<(ij)})\bigr)$ with
\begin{equation}\label{eq:quantization-step}
    q(\Wpi{ij}, \lambda, \gamma, P)
    {\,:\!=\,} \argmin_{g \in G} \biggl[ \frac12 \frac{(\Wpi{ij} - g)^2}{\Cpi{jj}^2} - \lambda \log_2 P(g) - \frac{\lambda\gamma}{2} g^2 \biggr]
\end{equation}

\begin{wrapfigure}{R}{0.58\textwidth}
\begin{minipage}{0.58\textwidth}
\vspace{-2.4em}
\begin{algorithm}[H]
    \caption{Compression with Entropy-Regularized Weight Updates (CERWU) (row-major variant).}
    \label{algo:cerwu}
    \begin{algorithmic}[1]
        \Require uncompressed weight matrix $\mW \in\mathbb R^{n\times m}$;
        \Statex grid size $k$ (see \autoref{eq:grid});
        \Statex Hessian $\mH = 2\mX \mX^T \in \mathbb R^{m\times m}$ (see \autoref{eq:renaming});
        \Statex rate/distortion trade-off parameter $\lambda>0$;
        \Statex autoreg.\ entropy model $P$ (e.g., DeepCABAC~\cite{wiedemannDeepCABACUniversalCompression2020}).
        \Ensure quantized weights $\Wh \in G^{n\times m}$ (see \autoref{eq:grid}) that can be well compressed with the entropy model~$P$.\vspace{0.3em}
        \State Set $\gamma \gets 1/(\ln(2) \operatorname{Var}(\{W_{ij}\}_{ij}))$ \Mycomment{See  \autoref{eq:regularized-ll}.}
        \State Set $\Hp \gets \mH + \lambda \gamma \mI$
        \State Initialize $\Wp \gets \mW \mH \left(\Hp\right)^{-1}$ \Mycomment{See \autoref{eq:renaming}.}

        \State Set $\Cp \gets \operatorname{Cholesky}(\left(\Hp\right)^{-1})^{T} $ \Mycomment{(upper triangular)}
        \State Initialize $P \gets \text{initEntropyModel}()$ \vspace{0.3em}
        \For{row $i = 1$ to $n$} 
            \For{column $j = 1$ to $m$} 
            \State Set $\Whi{ij} \gets q(\Wpi{ij},\lambda,\gamma,P)$ \label{line-quantize}
            \Mycomment{See \autoref{eq:quantization-step}.}
            \State Update $\mW_{i,>j}' \gets \mW_{i,>j}' - \frac{W'_{ij} - \Whi{ij}} {\Cpi{jj}} \mCpi{j,>j}$
            \Statex \Mycomment{See \autoref{eq:optq-correction}.}
            \State $P.\text{autoregressiveUpdate}(\Whi{ij})$
            \label{line-update-cabac}
            \EndFor
        \EndFor
    \end{algorithmic}\label{algo-optq-rd}
\end{algorithm}
\end{minipage}
\vspace{-2.5em}
\end{wrapfigure}

\paragraph{Scan order.}
When using an autoregressive entropy model, the order in which the weights are fed to the model starts to matter. There exist two different possibilities, \textit{row-major} (traversing along rows first) and \textit{column-major} (columns first). Although we did not observe a large performance difference between these two, we include both in our experiments and select the best version for each method. 

\paragraph{Complexity analysis.} Running the algorithm for a full network first requires a single forward pass over a moderately sized calibration set (e.g., we used 40'000 data points for ImageNet, around 3\% of the full training data set). 
The calculated activations can be cached for each layer, and future runs to quantize the network with different compression strengths~$\lambda$ do not need to repeat this forward pass. 

As for the quantization algorithm itself, assume a grid of size $k$. Each layer of size $n \times m$ requires $O(m^3)$ operations to calculate the Cholesky decomposition of the $m \times m$ sized inverse regularized Hessian~$(\Hp)^{-1}$. The quantization procedure requires $O(nm)$ quantization steps, each of which require an $O(k)$ grid search followed by an $O(m)$ row update. Therefore, the full complexity is $O(m^3 + nm^2 + nmk)$. Usually, it holds that $k \ll m$ for most layers, 
however, the grid search requires $O(k)$ calls to the entropy model. 
The cost for a single evaluation of the entropy model usually does not depend on $m,n,k$, but can still be expensive, depending on the chosen model.
We show empirical runtimes in \autoref{sec:runtimes}.

\section{Experiments}\label{sec:experiments}

\paragraph{Experiment setup.}
We test our proposed quantization method on multiple networks from the computer vision community. We include ResNet-\{\!\!\;18,\,52,\,101,\,152\}~\citep{heDeepResidualLearning2016}, VGG16~\citep{simonyanVeryDeepConvolutional2015} and MobileNetv3-\{\!\!\;small,\,large\}~\citep{howardSearchingMobileNetV32019} trained on ImageNet~\citep{deng2009imagenet}, as well as ResNet-\{\!\!\;18,\,34,\,52\} trained on CIFAR10~\citep{krizhevskyCIFAR10CanadianInstitute}. 
For our entropy model, we use DeepCABAC~\cite{wiedemannDeepCABACUniversalCompression2020} due to its high speed and its focus on
modeling typical neural network weight statistics. For all methods, we perform a sweep over the scan-order, various compression strength parameters and grid sizes to ensure a good coverage of the rate-distortion curve.
Shown bit rates are the measured bit rates after actual entropy coding plus a small overhead for storing the grid spacings for each layer.
Refer to \autoref{app:experimental} for more detailed information.

\paragraph{Pareto front.} To visualize the results from our search space, we calculate a Pareto front in the rate-distortion space in the following manner: for a given network and compression method, we iterate over all 
data points $(\text{rate}, \text{accuracy})$ obtained from our sweep. For each data point $x$, if there exists a point $y$ that has both a lower rate as well as higher or equal accuracy, we discard~$x$.

\paragraph{Ablations and baselines.}
Aside from our main proposed compression algorithm detailed in \algoref{algo-optq-rd}, we also explore two variations: a version with unregularized weight updates, which amounts to artificially setting $\gamma = 0$, and a version that does not account for the rate during quantization, which amounts to setting $\lambda = 0$ (this also renders the method independent of~$\gamma$). The case of $\lambda = 0$ can also be interpreted as performing OPTQ \citep{frantar2023optq} followed by entropy coding with the chosen entropy model. These two methods are labeled as $\mathbf{CERWU\text{-}\boldsymbol\gamma\!=\!0}$ and $\mathbf{CERWU\text{-}\boldsymbol\lambda\!=\!0}$ respectively. 
We also compare our method against the state-of-the-art compression scheme \textbf{NNCodec} \cite{beckingNNCodecOpenSource2023}, an implementation of the Neural Network Compression Standard ISO/IEC 15938-17 also based on the DeepCABAC entropy model. As a further baseline, we include a simple quantization scheme of nearest-neighbour quantization followed with entropy coding via DeepCABAC, which we label as \textbf{RTN+EC}.

\subsection{Compression Performance}

\begin{figure}[h]
    \centering
    \includegraphics[width=\textwidth]{"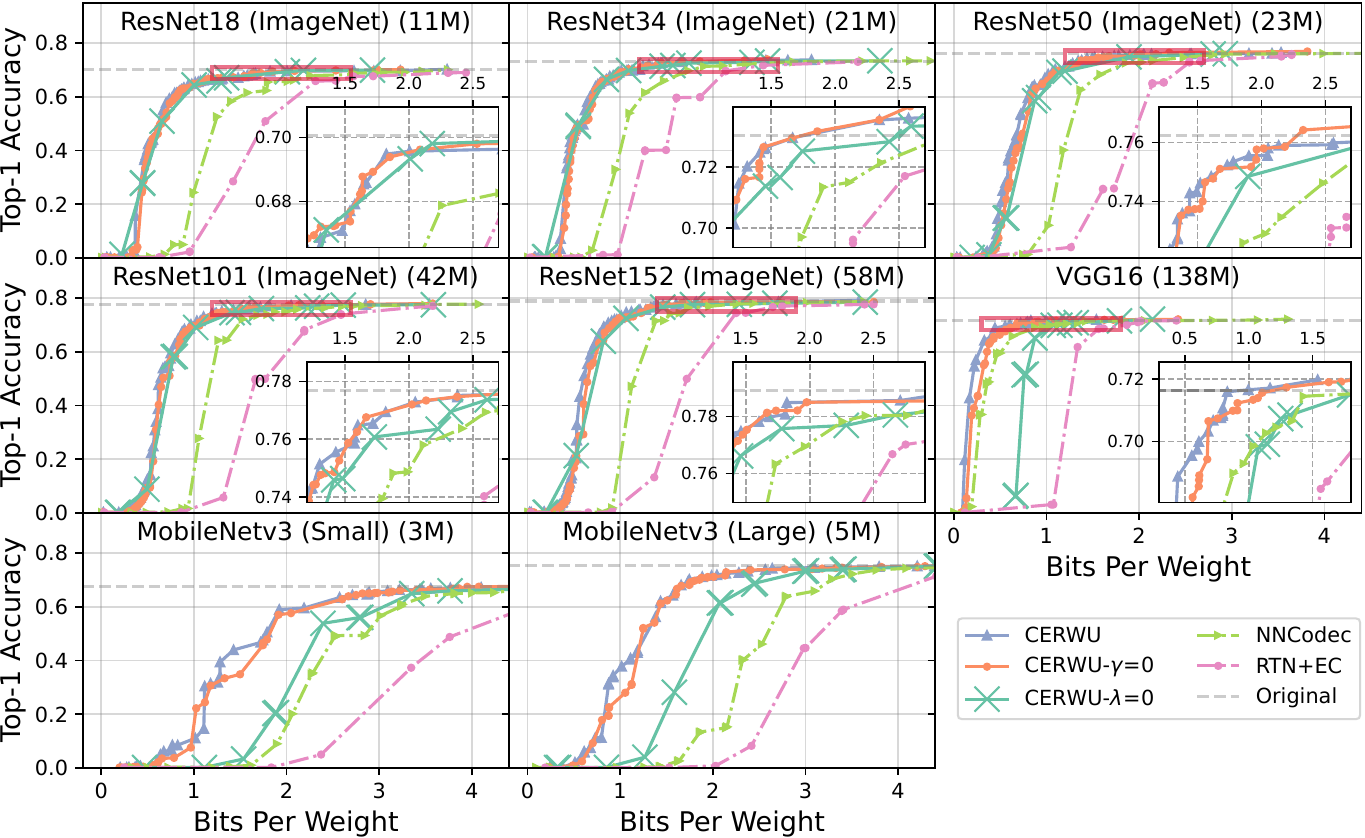"}
    \caption{Performance of our compression methods on various networks trained and evaluated on ImageNet. For better visibility, a Pareto front over the parameters was calculated for each curve. The inset at the bottom right of each plot shows a zoomed-in version of the area marked in the red bounding box. The inset ranges over $(0.95 \cdot \text{acc}_{\text{orig.}}, 1.0125 \cdot \text{acc}_{\text{orig.}})$ on the y-axis and encompasses a 1.5-bits-per-weight range in the x-axis. Our proposed methods and ablations are marked with solid lines, 
    baselines are marked in dashed lines, and the original performance of the (uncompressed) network is marked with a horizontal, gray dashed line. The plot titles include the number of quantizable
    parameters each network has (multiply by 32 bit to get the uncompressed storage size on disk).}
    \label{fig:lines-imagenet}
\end{figure}

\begin{figure}[h]
    \centering
    \includegraphics[width=\textwidth]{"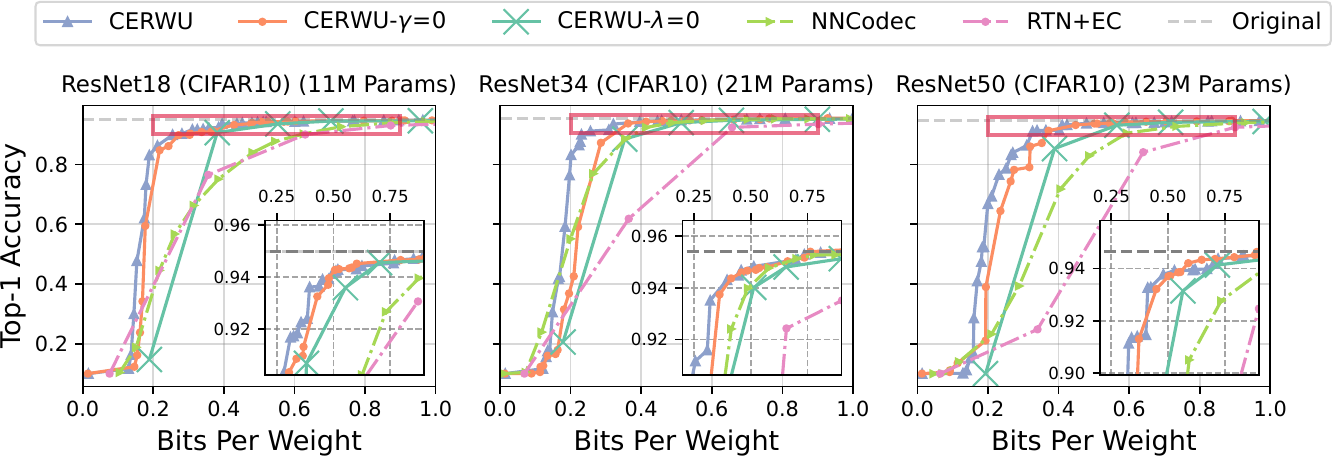"}
    \caption{Performance of compression methods for CIFAR10-trained networks, analogous to \autoref{fig:lines-imagenet}.}
    \label{fig:lines-cifar10}
\end{figure}

\begin{figure}[h]
    \centering
    \includegraphics[width=\textwidth]{"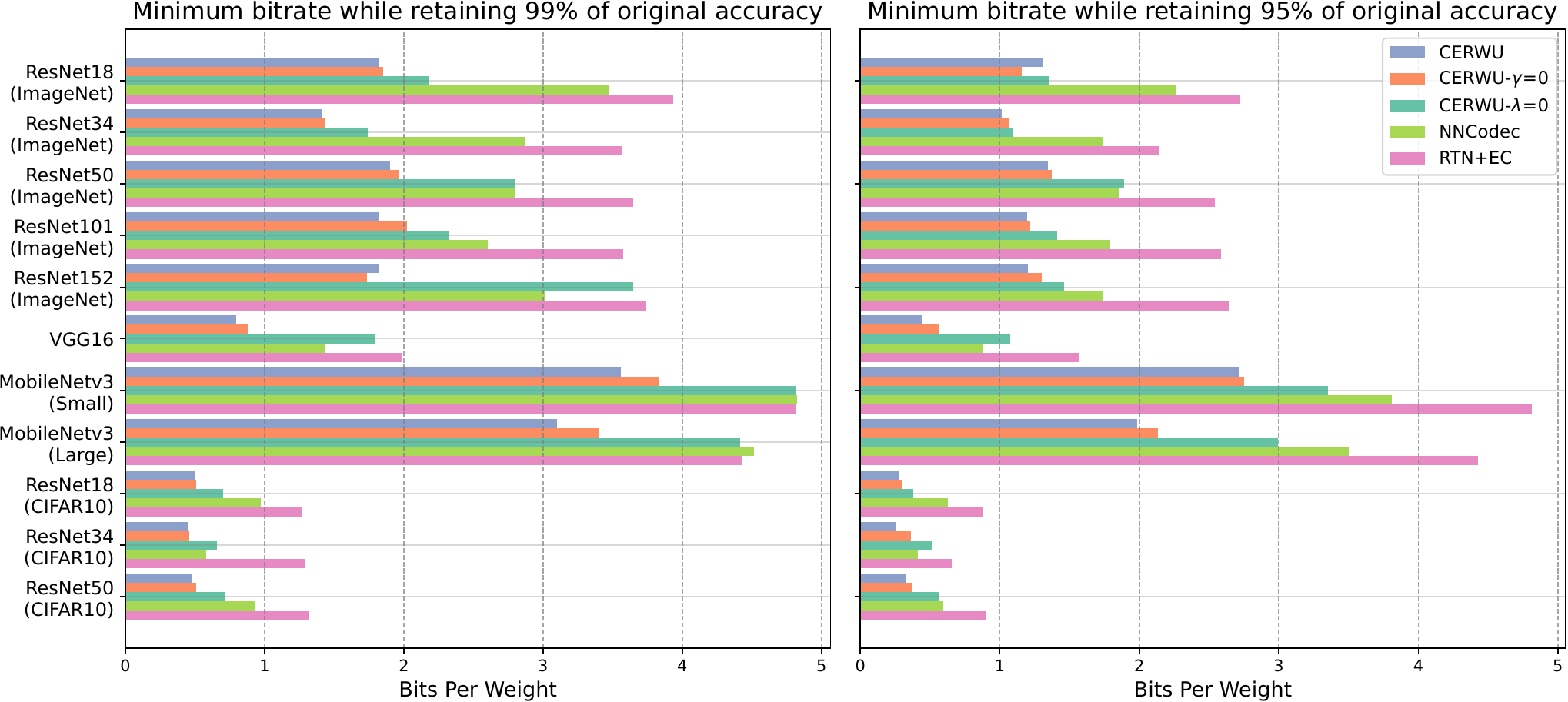"}
    \caption{Minimum bits per weight achieved at 99\% (left) and 95\% (right) of the original test accuracy for different methods.
    Lower is better.
    Both CERWU (blue) and CERWU-$\gamma\!=\!0$ (orange) outperform all other methods, with CERWU achieving a slight edge over CERWU-$\gamma\!=\!0$ for most networks.}
    \label{fig:bar-performance}
\end{figure}

\autoref{fig:lines-imagenet} presents a comparison of the three variants of our proposed compression method and the baselines on our analyzed networks for ImageNet.
\autoref{fig:lines-cifar10} shows the same for networks trained on CIFAR10.
Additionally, \autoref{fig:bar-performance} shows bar plots of the bit rate achieved by each method for the strongest compression setting that still retains $99\%$ and $95\%$, respectively, of the original performance. 

We observe that CERWU and CERWU-$\gamma\!=\!0$ consistently 
outperform all other tested methods, achieving a more favorable rate-distortion performance over most of the rate-distortion curve. 
While in the line plots it can be hard 
to distinguish between CERWU and CERWU-$\gamma\!=\!0$, the bar plots can help distinguish
between them.
There, we see that CERWU maintains a slight but significant edge (1-5\% stronger compression performance) for most networks for both evaluated fixed performance levels over CERWU-$\gamma\!=\!0$. This demonstrates that taking estimations of
future rates into account when performing the weight updates helps to create more strongly compressible representations.

Compared to the baselines, CERWU and CERWU-$\gamma\!=\!0$
often achieve much stronger compression rates when compressing the network
to almost-original performance (which we expect to be the regime of interest for most practical applications), producing a compressed representation that is 20-40\% more strongly compressed
than the current compression standard NNCodec, see \autoref{fig:bar-performance}.

\subsection{Run times}\label{sec:runtimes}
\begin{figure}[h]
    \centering
    \includegraphics[width=1.0\textwidth]{"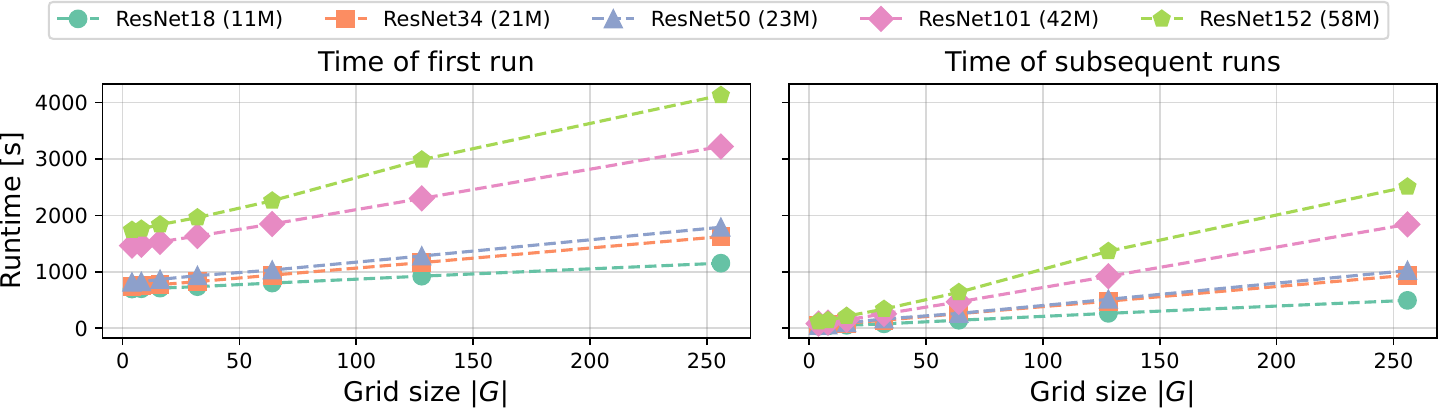"}
    \caption{Run times for compressing ResNets of differing sizes. Left: run times for the first run; Right: run times for subsequent runs for different values of~$\lambda$ or~$k$ (which can reuse the Hessian).}
    \label{fig:runtimes}
\end{figure}

\paragraph{Encoding times.} We timed our proposed compression method CERWU on our hardware (described in \autoref{app:experimental}) for ImageNet-trained ResNets of varios sizes (the size of the training set does not influence encoding times).
Initial runs (\autoref{fig:runtimes}~left) include the time for the forward passes to calculate the Hessian.
Subsequent runs (\autoref{fig:runtimes}~right) are considerably faster as they can reuse the Hessian~$\mH$ for different values of~$\lambda$ or~$k$.
Especially in the regime of 
small grid sizes~$k$ ($\le 31$, which are usually sufficient to produce
high-accuracy compressed representations), the run time is heavily
dominated by the calculation of the Hessian, and runs after the first one become quite cheap. 
The baselines NNCodec and RTN+EC only require between 1-10 seconds to quantize each network. This is much faster than our method for large grids or for the first run.
While our achieved encoding times are still in the manageable realm, 
further work to speed up the grid search might be helpful (such as introducing early stopping), as well as a thorough optimization of 
our implementation.

\paragraph{Decoding times.} The \textit{decoding time} is the more important metric, as often compression is performed once on a device with larger computational resources, while decompression 
is performed many times on weaker devices. Here, by
construction of our method, we achieve the same decompression 
speed as NNCodec or RTN+EC, since we can decompress our network using the same decoder in form of the DeepCABAC entropy model (the decompression algorithm is oblivious to the distortion function and quantization scheme used during compression).
The decompression speed is between 0.2 and 1.2~seconds for all networks and all tested methods on our hardware (see \autoref{app:experimental}) despite using a single-threaded decoder implementation.

\section{Conclusion}\label{sec:conclusion}
We showcased a post-training compression method that achieves state-of-the-art storage sizes for computer vision architectures suitable for resource constrained devices. We showed how adding a quadratic rate estimation to the layerwise loss provides locally exact solutions to the resulting rate/distortion constrained weight update problem, extending the OBS framework on which many contemporary compression works build~\citep{frantarOptimalBrainCompression2023,frantar2023optq,frantarSparseGPTMassiveLanguage2023}. Our experiments demonstrate a 20-40\% reduction in file size at no additional accuracy drop compared to the state-of-the-art method NNCodec.

Moreover, the flexibility of our method opens up interesting avenues for future research directions and applications. 
For example, our method can be used with different entropy models and quantization grids, and the resulting compression performance can be evaluated.
Additionally, given a suitable entropy model, 
our method can also produce compressible representations that
could be decoded \textit{on-the-fly} on a GPU. This might serve to reduce
the required communication bandwidth between RAM and GPU during 
inference, resulting in possible energy savings and improved latency.

\paragraph{Limitations.} While our work shows strong compression performance on (convolutional) computer vision architectures, we do not include an analysis for transformer-based models. We argue that resource-constrained devices often resort to simpler architectures
such as those that we surveyed due to their weaker computational power. 
In particular, we do not discuss (large) language models, where some quick experiments (see \autoref{app:pythia}) indicate worse performance of both our method and NNCodec, but where we argue that there already exists a large body of work that specifically targets these models \citep{vanbaalenGPTVQBlessingDimensionality2024,egiazarianExtremeCompressionLarge2024a,sunSimpleEffectivePruning2024,dongSTBLLMBreaking1Bit2024,ashkboosQuaRotOutlierFree4Bit2024}, often taking specific
architecture-dependent behavior into account (e.g., outliers in the channel activations \citep{linAWQActivationawareWeight2024,kimSqueezeLLMDenseandSparseQuantization2024}). 

\FloatBarrier

\section*{Acknowledgements}

We thank Tim Z.\ Xiao for pointing us to relevant literature, and Julia Dietl for insightful discussions and an early exploration of model compression methods.

This work was partially funded by the Deutsche Forschungsgemeinschaft (DFG, German Research Foundation) under Germany's Excellence Strategy – EXC number 2064/1 – Project number 390727645. 
This work was supported by the German Federal Ministry of Education and Research (BMBF): Tübingen AI Center, FKZ: 01IS18039A.
RB acknowledges funding by the German Research Foundation (DFG) for project 448588364 of the Emmy Noether Program.

\bibliographystyle{plain} 
\bibliography{neurips_2025.bib}


\newpage
\appendix
\section{Experimental Details}\label{app:experimental}

\paragraph{Evaluation details.}
For the tested networks, we report the accuracy, which is the top-1 accuracy and the average number of bits required to encode one weight. 
The bits per weight are obtained by dividing the total compressed size by the number of compressed parameters, while 
the total compressed size consists of the size of the entropy coded weights plus the overhead of one 16-bit scale parameter per tensor. For ResNets, we don't quantize the last layer, as well as the BatchNorm layers, as they can often be fused together with preceding layers. To ensure a fair comparison, these layers are not quantized for all analyzed methods. To calculate
the layer-wise loss for convolutional layers, we unfold convolutional layers (see f.e. \citep{nagelAdaptiveRoundingPostTraining2020,hubaraAccuratePostTraining2021,frantarOptimalBrainCompression2023})
For the MobileNet networks to be compressed with our method, we pre-process the grouped convolutions to
be represented as $n_{\text{groupsize}}$ single-filter convolutions, whose results are then concatenated (this does not change the network output). We then compress each convolution separately, taking care to count one scale parameter per filter to the total compression overhead.

\paragraph{Parameter choices.} 
To calculate the hessians, we use calibration samples from the training set containing 40'000 images for ImageNet and 64'000 images for CIFAR10. For ImageNet, we use 16'000 samples from the validation set to calculate the top-1 accuracy, and for CIFAR10, we use 10'000 samples from the test set. 
For the grid, we use a symmetric uniform grid and sweep between a subset of grid sizes \{4,6,8,12,16,32,48,64,128,256,512,1024\}, 
depending on the method and network. For the methods that include a $\lambda$ parameter, we perform a sweep between values of $10^{-8}$ and $10^{-1}$ at steps of $0.5$ in log-space, with increased granularity of $0.1$ in regions of high variability. Additionally, for every method, we once iterate over the weights in row-major and once in column-major order. For NNCodec, we swept over the qp parameter in a $(-38,-4)$ interval.

\paragraph{Soft- and hardware.} Our algorithm is implemented mainly in C++, while the code associated with model loading and evaluation was implemented using the PyTorch library \cite{paszkePyTorchImperativeStyle2019}, whose pre-trained models for ImageNet were used for our evaluations. 
The pre-trained ResNet models for CIFAR10 were obtained from a public repository of user edaltocg on HuggingFace.
The quantization part of the algorithm was executed on a 2.6 GHz Intel Xeon Gold 6240, while the evaluation and calculation of the Hessian were done on a single 
NVIDIA 2080ti graphics card. 

\section{Pythia Evaluations.}\label{app:pythia}

\begin{figure}[h]
    \centering
    \includegraphics[width=0.9\textwidth]{"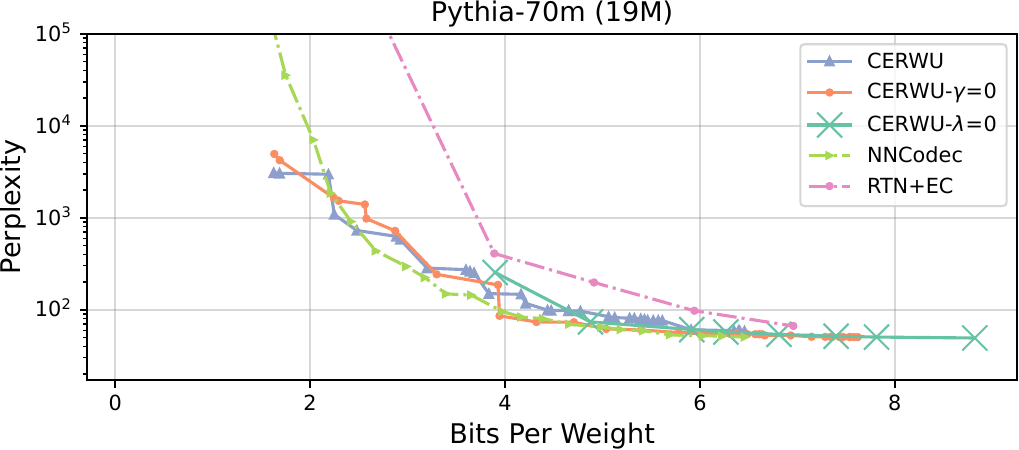"}
    \caption{Rate-distortion performance of our methods on the small language model Pythia-70M. We swept over grid sizes of \{4, 16, 128, 256, 352, 512, 768, 1024, 2048\}.}
    \label{fig:pythia}
\end{figure}

\section{Derivation of the Entropy-Regularized Weight Update}\label{app:derivation}

This section provides a derivation of Equations~\ref{eq:obs-update} and~\ref{eq:optq-correction} in compact form.
We stress that the derivation is not new, but rather combines arguments by Hassibi et al.\ in~\cite{hassibiOptimalBrainSurgeon1993} and by Frantar et al.\ in~\cite{frantar2023optq}, applying them to the quadratic part $L'_\lambda(\Wh)$ of our entropy-regularized layerwise loss function in \autoref{eq:regularized-ll}.
We provide the derivation here for completeness, to clarify what it means precisely when we say that the weight updates ``optimally compensate'' for introduced rounding errors, to show how analogous arguments as in~\cite{hassibiOptimalBrainSurgeon1993,frantar2023optq} apply to our modified loss function, and to motivate why the weight updates in our proposed CERWU method take the rate term into account only within a quadratic approximation (as opposed to the quantization in \autoref{eq:quantization-step}, which also takes the non-quadratic part $L''_\lambda(\Wh)$ into account).

\subsection{Verification of \autoref{eq:regularized-ll-simple}}\label{app:completing-the-square}
Before we formalize what we mean with ``optimal weight updates'', we first verify that \autoref{eq:regularized-ll-simple} is indeed equivalent to the definition of $L'_\lambda(\Wh)$ in \autoref{eq:regularized-ll}.
Using the squared Frobenius norm $\|\mA\|_2^2 {\,:\!=\,} \sum_{i,j}A_{ij}^2 = \Tr[\mA \mA^T]$ for any matrix~$\mA$, and $\mH {\,:\!=\,} 2\mX\mX^T$, we multiply out the right-hand side of \autoref{eq:regularized-ll} and find,
\begin{align}\label{eq:expand-regularized-ll}
    \text{r.h.s.\ of \autoref{eq:regularized-ll}}
    &= \| (\mW - \Wh) \mX \|^2_2 + \frac{\lambda\gamma}{2} \|\Wh\|_2^2 \nonumber\\
    &= \frac12 \Tr\Bigl[ \bigl(\mW - \Wh\bigr) \mH \bigl(\mW - \Wh\bigr)^T + \lambda\gamma \Wh \Wh^T \Bigr] \nonumber\\
    &= \frac12 \Tr\Bigl[
        \underbrace{\mW\mH\mW^T}_\text{const.}
        \,-\; 2\mW\mH\Wh^T + \Wh\mH\Wh^T + \lambda\gamma \Wh \Wh^T
        \Bigr]
\end{align}
where the ``const.'' term is independent of $\Wh$.
By comparison, inserting $\Hp = \mH + \lambda\gamma \mI$ and $\Wp = \mW\mH(\Hp)^{-1}$ from \autoref{eq:renaming} into \autoref{eq:regularized-ll-simple}, we find (using that $\mH$ and $\Hp$ are symmetric):
\begin{align}\label{eq:expand-regularized-ll-simple}
    \text{r.h.s.\ of \autoref{eq:regularized-ll-simple}}
    &= \frac12 \Tr\Bigl[ (\Wp - \Wh) \Hp (\Wp - \Wh)^T \Bigr] + \text{const.} \nonumber\\
    &= \frac12 \Tr\Bigl[ \Bigl(\mW\mH(\Hp)^{-1} - \Wh\Bigr) \Hp \Bigl((\Hp)^{-1}\mH\mW^T - \Wh^T \Bigr) \Bigr] + \text{const.} \nonumber\\
    &= \frac12 \Tr\Bigl[
        \underbrace{\mW \mH (\Hp)^{-1} \mH \mW^T}_\text{const.\ (independent of $\Wh$)}
        \,-\; 2\mW \mH \Wh^T + \Wh \Hp \Wh^T
        \Bigr] + \text{const.} \nonumber\\
    &= \frac12 \Tr\Bigl[
        - 2\mW \mH \Wh^T + \Wh (\mH + \lambda\gamma\mI) \Wh^T
        \Bigr] + \text{const.}
\end{align}
Thus, Equations~\ref{eq:regularized-ll} and~\ref{eq:regularized-ll-simple} are indeed equivalent.
The advantage of \autoref{eq:regularized-ll-simple} is that its simpler form allows us to immediately see that the Hessian of $L'(\Wh)$ is~$\Hp$, and that $L'(\Wh)$ would be minimized if we could set~$\Wh$ to the entropy regularized weight matrix $\Wp = \mW\mH(\Hp)^{-1}$ (which, however, is not possible because $\Wp \notin G^{n\times m}$ in general).

\subsection{Formalization of the Propositions}\label{app:propositions}

For $\Wp \in \mathbb R^{n\times m}$ and $\Wh\in G^{n\times m}$, the quadratic loss $L'_\lambda(\Wh)$ in \autoref{eq:regularized-ll-simple} separates over the rows,
\begin{align}\label{eq:separate-rows}
    L'_\lambda(\Wh) = \sum_{i=1}^n L'_{i,\lambda}(\Wh_{i,:}) + \text{const.},
    \quad
    L'_{i,\lambda}(\Wh_{i,:}) = \frac12 (\mW'_{i,:} - \Wh_{i,:}) \Hp (\mW'_{i,:} - \Wh_{i,:})^T
\end{align}
where the notation $\mA_{i,:}$ denotes a row vector comprised of the $i$-th row of a matrix~$\mA$, and $\Hp\in\mathbb R^{m\times m}$ (see \autoref{eq:renaming}) is symmetric and positive definite.
Due to the separation over rows, minimizing $L'_\lambda(\Wh)$ over~$\Wh$ is equivalent to independently minimizing $L'_{i,\lambda}(\Wh_{i,:})$ for each row $i\in\{1,\ldots,n\}$, and we therefore focus the remaining discussion on a given row~$i$.
Our goal is to show that the weight updates in \autoref{eq:obs-update} and \autoref{eq:optq-correction} are equivalent, and that they optimally compensate subsequent weights in the same row~$i$ for the error introduced by quantizing $W'_{ij}$ to~$\Whi{ij}$.

More formally, consider starting with the unquantized row $\mW^{\prime(0)}_{i,:} {\,:\!=\,} \mW'_{i,:} \in \mathbb R^{1\times m}$, and iterating over its entries $j\in\{1,\ldots,m\}$ in ascending order.
At each iteration~$j$, we construct an updated row vector $\mW^{\prime(j)}_{i,:} \in \mathbb R^{1\times m}$ by leaving the first $j-1$ entries unchanged (i.e., $\mW^{\prime(j)}_{i,<k} {\,:\!=\,} \mW^{\prime(j-1)}_{i,<k} = \Wh_{i,<k}$), quantizing the $j$-th entry $W^{\prime(j)}_{ij} {\,:\!=\,} \Whi{ij}$ to some (for the sake of the proof below) arbitrary value~$\Whi{ij}$, and setting the remaining $m-j$ entries to
\begin{align}\label{eq:claim-formal}
    \mW^{\prime(j)}_{i,>j} {\,:\!=\,} \argmin_{\bm{\tilde W}_{i,>j} \in \mathbb R^{1\times (m-j)}}
    L'_{i,\lambda}(\bm{\tilde W}_{i,:})
    \qquad\text{with the constraint}\qquad
    \bm{\tilde W}_{i,\leq j} = \Wh_{i,\leq j},
\end{align}
where we formally extend the domain of $L'_{i,\lambda}$ from $G^{1\times m}$ to $\mathbb R^{1\times m}$ according to \autoref{eq:separate-rows}.

\paragraph{Propositions.}
For $j\in\{1,\ldots,m\}$ and $\mW^{\prime(j)}_{i,>j}$ given in \autoref{eq:claim-formal}, we claim that
\begin{enumerate}
    \item[(i)]
        $\mW^{\prime(j)}_{i,>j} = \mW^{\prime(j-1)}_{i,>j} + \Delta_{\mW'_{i,>j}}$ where $\Delta_{\mW'_{i,>j}}$ is given in \autoref{eq:obs-update};
    \item[(ii)]
        the quantization and subsequent weight update increase the quadratic loss by
        \begin{align}
            L'_{i,\lambda}\bigl(\mW^{\prime(j)}_{i,:}\bigr) - L'_{i,\lambda}\bigl(\mW^{\prime(j-1)}_{i,:}\bigr) = \Delta_{L'}
        \end{align}
        where $\Delta_{L'}$ is also given in \autoref{eq:obs-update}; and
    \item[(iii)]
        Equations~\ref{eq:obs-update} and~\ref{eq:optq-correction} are equivalent expressions for $\Delta_{\mW'_{i,>j}}$ and $\Delta_{L'}$.
\end{enumerate}

\subsection{Proof of Proposition (i)}

We prove proposition~(i) by induction over $j\in\{0,\ldots,m\}$.
For $j=0$, we already defined $\mW^{\prime(0)}_{i,:} {\,:\!=\,} \mW'_{i,:}$, which is trivially equivalent to \autoref{eq:claim-formal} because $L'_{i,\lambda}$ is positive definite with its minimum at $L'_{i,\lambda}(\mW'_{i,:}) = 0$.
Consider now some $j\in\{1,\ldots,m\}$ and assume that $\mW^{\prime(j-1)}_{i,< k} = \Wh_{i,< k}$, and that \autoref{eq:claim-formal} holds for $j-1$, i.e.,
\begin{align}
    \mW^{\prime(j-1)}_{i,\geq j} &= \argmin_{\bm{\tilde W}_{i,\geq j} \in \mathbb R^{1\times (m-j+1)}}
    L'_{i,\lambda}(\bm{\tilde W}_{i,:})
    \qquad\text{with the constraint}\qquad
    \bm{\tilde W}_{i,<j} = \Wh_{i,<j} \label{eq:induction-hypothesis}
\end{align}
(note the ``$\geq$'' instead of ``$>$'', and ``$<$'' instead of ``$\leq$'' when comparing \autoref{eq:induction-hypothesis} to \autoref{eq:claim-formal}).
We simplify \autoref{eq:induction-hypothesis} by splitting $L'_{i,\lambda}(\bm{\tilde W}_{i,:})$ from \autoref{eq:separate-rows} into four parts:
\begin{align}
    L'_{i,\lambda}(\bm{\tilde W}_{i,:})
    &= \frac12 \bigl(
            \mW^{\prime(0)}_{i,<j} - \bm{\tilde W}_{i,<j} ,\;
            \mW^{\prime(0)}_{i,\geq j} - \bm{\tilde W}_{i,\geq j}
        \bigr)
        \begin{pmatrix}
            \mH'_{<j,<j} & \mH'_{<j,\geq j} \\
            \mH'_{\geq j,<j} & \mH'_{\geq j,\geq j}
        \end{pmatrix}
        \begin{pmatrix}
            \bigl(\mW^{\prime(0)}_{i,<j} - \bm{\tilde W}_{i,<j} \bigr)^{\!T\,} \\
            \bigl(\mW^{\prime(0)}_{i,\geq j} - \bm{\tilde W}_{i,\geq j} \bigr)^{\!T\,}
        \end{pmatrix}
\end{align}
where we wrote $\mW'$ explicitly as $\mW^{\prime(0)}$ to stress that $L'_{i,\lambda}$ always compares to the \emph{original} unquantized row, not to any updated version.
By \autoref{eq:induction-hypothesis}, we know that $\mW^{\prime(j-1)}_{i,:}$ is a stationary point of $L'_{i,\lambda}(\bm{\tilde W}_{i,:})$ with respect to $\bm{\tilde W}_{i,\geq j}$ under the constraint $\bm{\tilde W}_{i,<j} = \Wh_{i,<j} \,(= \mW^{\prime(j-1)}_{i,<j})$, i.e.,
\begin{align}
    \bm{0}_{1\times m-j+1}
    &= \nabla_{\!\bm{\tilde W}_{i,\geq j}} L'_{i,\lambda}(\bm{\tilde W}_{i,:})
        \big|_{\bm{\tilde W}_{i,:} = \mW^{\prime(j-1)}_{i,:}} \nonumber\\
    &= - \Bigl(\mW^{\prime(0)}_{i,<j} - \underbrace{\mW^{\prime(j-1)}_{i,<j}}_{=\Wh_{i,<j}} \Bigr) \mH'_{<j,\geq j}
        - \Bigl(\mW^{\prime(0)}_{i,\geq j} - \mW^{\prime(j-1)}_{i,\geq j}\Bigr) \mH'_{\geq j,\geq j} \nonumber\\
    &= \Wh_{i,<j} \mH'_{<j,\geq j}
       + \mW^{\prime(j-1)}_{i,\geq j} \mH'_{\geq j,\geq j}
       - \mW^{\prime(0)}_{i,:} \mH'_{:,\geq j}.
       \label{eq:stationary1}
\end{align}

Similarly, by definition in \autoref{eq:claim-formal}, $\mW^{\prime(j)}_{i,:}$ is also a stationary point of $L'_{i,\lambda}(\bm{\tilde W}_{i,:})$, however, this time only with respect to $\bm{\tilde W}_{i,>j}$ rather than $\bm{\tilde W}_{i,\geq j}$, i.e., the derivative with respect to $\tilde W_{ij}$ at $\mW^{\prime(j)}_{i,:}$ may take some (in general) non-zero value~$\gamma_{ij}$.
It turns out that the following calculations are easier if we nevertheless include the derivative with respect to $\tilde W_{ij}$, and explicitly keep track of the fact that it is not necessarily zero.
Thus,
\begin{align}
    (\gamma_{ij},\; \bm{0}_{1\times m-j})
    &= \nabla_{\!\bm{\tilde W}_{i,\geq j}} L'_{i,\lambda}(\bm{\tilde W}_{i,:})
        \big|_{\bm{\tilde W}_{i,:} = \mW^{\prime(j)}_{i,:}} \nonumber\\
    &= \Wh_{i,<j} \mH'_{<j,\geq j}
       + \mW^{\prime(j)}_{i,\geq j} \mH'_{\geq j,\geq j}
       - \mW^{\prime(0)}_{i,:} \mH'_{:,\geq j}.
       \label{eq:stationary2}
\end{align}
Subtracting \autoref{eq:stationary1} from \autoref{eq:stationary2} leads to a cancellation of the terms with $\Wh_{i,<j}$ and $\mW^{\prime(0)}_{i,:}$, and we find
\begin{align}
    \mW^{\prime(j)}_{i,\geq j} \mH'_{\geq j,\geq j}
    &= \mW^{\prime(j-1)}_{i,\geq j} \mH'_{\geq j,\geq j} + (\gamma_{ij},\; \bm{0}_{1\times m-j}).
\end{align}
We solve for $\mW^{\prime(j)}_{i,\geq j}$ by multiplying from the right with $(\mH'_{\geq j,\geq j})^{-1}$, and find
\begin{align}\label{eq:proposition1-with-gamma}
    \mW^{\prime(j)}_{i,\geq j}
    &= \mW^{\prime(j-1)}_{i,\geq j} + (\gamma_{ij},\; \bm{0}_{1\times m-j}) (\mH'_{\geq j,\geq j})^{-1}
    = \mW^{\prime(j-1)}_{i,\geq j} + \gamma_{ij} \bigl[(\mH'_{\geq j,\geq j})^{-1}\bigr]_{j,\geq j}.
\end{align}
Finally, we solve for $\gamma_{ij}$ by evaluating \autoref{eq:proposition1-with-gamma} at indices $(i,j)$, recalling that $W^{\prime(j)}_{ij} = \Whi{ij}$ by definition.
Thus,
\begin{align}
    \Whi{ij}
    &= W^{\prime(j-1)}_{ij} + \gamma_{ij} \bigl[(\mH'_{\geq j,\geq j})^{-1}\bigr]_{jj}
    \qquad\Longrightarrow\qquad
    \gamma_{ij} = \frac{\Whi{ij} - W^{\prime(j-1)}_{ij}}{\bigl[(\mH'_{\geq j,\geq j})^{-1}\bigr]_{jj}}.
\end{align}
Inserting this result for $\gamma_{ij}$ into \autoref{eq:proposition1-with-gamma} leads to $\mW^{\prime(j)}_{i,>j} = \mW^{\prime(j-1)}_{i,>j} + \Delta_{\mW'_{i,>j}}$ with $\Delta_{\mW'_{i,>j}}$ as given in \autoref{eq:obs-update} of the main text (recall that, in \autoref{eq:obs-update}, $\Wpi{ij}$ refers to an unquantized weight that has already been updated in the previous iterations, see Algorithm~\ref{algo:cerwu}; this is $W^{\prime(j-1)}_{ij}$ in the more explicit notation used in this appendix).

\subsection{Proof of Proposition (ii)}

For fixed $\Wh_{i,<j}$, we consider the function $\bm{\tilde W}_{i,\geq j} \mapsto L'_{i,\lambda}\bigl((\Wh_{i,<j},\; \bm{\tilde W}_{i,\geq j})\bigr)$.
Since this is a quadratic function with Hessian $\mH'_{\geq j,\geq j}$ and a stationary point at $\mW^{\prime(j-1)}_{i,\geq j}$ by \autoref{eq:induction-hypothesis}, we have
\begin{align}\label{eq:proposition-ii-derivation}
\begin{split}
    L'_{i,\lambda}\bigl((\Wh_{i,<j},\; \bm{\tilde W}_{i,\geq j})\bigr)
    =&\, L'_{i,\lambda}\bigl((\Wh_{i,<j},\; \mW^{\prime(j-1)}_{i,\geq j})\bigr) \\
    &+ \frac12 \Bigl(\bm{\tilde W}_{i,\geq j} - \mW^{\prime(j-1)}_{i,\geq j} \Bigr)
        \mH'_{\geq j,\geq j} \Bigl(\bm{\tilde W}_{i,\geq j} - \mW^{\prime(j-1)}_{i,\geq j}\Bigr)^{\!T}.
\end{split}
\end{align}
Evaluating \autoref{eq:proposition-ii-derivation} at $\bm{\tilde W}_{i,\geq j} = \mW^{\prime(j)}_{i,\geq j}$ and subtracting the first term on the right-hand side of \autoref{eq:proposition-ii-derivation}, we thus find
\begin{align}
\begin{split}
    \Delta_{L'}
    &= L'_{i,\lambda}\bigl(\mW^{\prime(j)}_{i,:}\bigr) - L'_{i,\lambda}\bigl(\mW^{\prime(j-1)}_{i,:}\bigr) \\
    &= L'_{i,\lambda}\bigl((\Wh_{i,<j},\; \mW^{\prime(j)}_{i,\geq j})\bigr) - L'_{i,\lambda}\bigl((\Wh_{i,<j},\; \mW^{\prime(j-1)}_{i,\geq j})\bigr) \\
    &= \frac12 \Bigl(\mW^{\prime(j)}_{i,\geq j} - \mW^{\prime(j-1)}_{i,\geq j} \Bigr)
    \mH'_{\geq j,\geq j} \Bigl(\mW^{\prime(j)}_{i,\geq j} - \mW^{\prime(j-1)}_{i,\geq j}\Bigr)^{\!T} \\
    &= \frac12 \Delta_{\mW'_{i,>j}} \mH'_{\geq j,\geq j} (\Delta_{\mW'_{i,>j}})^T
\end{split}
\end{align}
where we used that $\mW^{\prime(j)}_{i,>j} - \mW^{\prime(j-1)}_{i,>j} = \Delta_{\mW'_{i,>j}}$.
Inserting $\Delta_{\mW'_{i,>j}}$ from \autoref{eq:obs-update}, we find
\begin{align}
    \Delta_{L'}
    &= \frac12 \left(\frac{\Wpi{ij} - \Whi{i,j}}{\Hipi{j}{jj}}\right)^{\!2}
        \underbrace{\Hipi{j}{j,>j} \mH'_{\geq j,\geq j} \Hipi{j}{>j,j}}_{=\Hipi{j}{jj}}
    = \frac12 \frac{(\Wpi{ij} - \Whi{i,j})^{2}}{\Hipi{j}{jj}}
\end{align}
as claimed in the second part of \autoref{eq:obs-update}.

\subsection{Proof of Proposition (iii)}
\label{app:proposition-iii}

To proof equivalence of \autoref{eq:obs-update} and \autoref{eq:optq-correction}, we have to show that, for all $k \geq j$,
\begin{align}\label{eq:proposition-iii-reduced-claim}
    \frac{\Hipi{j}{jk}}{\Hipi{j}{jj}} = \frac{\Cpi{jk}}{\Cpi{jj}}
    \qquad\text{and}\qquad
    \Hipi{j}{jj} = (C'_{jj})^2
\end{align}
where $\Cp$ is an upper triangular matrix that satisfies $\Cp^T \Cp = (\Hp)^{-1}$.
Using block-matrix notation,
\begin{align}
    \Cp = \begin{pmatrix}
        \mC'_{<j,<j} & \mC'_{<j,\geq j} \\
        \bf{0} & \mC'_{\geq j,\geq j}
    \end{pmatrix},
\end{align}
it is easy to verify (by multiplying out $(\Cp)^{-1} \Cp$ and verifying that the result is the identity) that
\begin{align}
    (\Cp)^{-1} = \begin{pmatrix}
        (\mC'_{<j,<j})^{-1} & -(\mC'_{<j,<j})^{-1} \,\mC'_{<j,\geq j}\, (\mC'_{\geq j,\geq j})^{-1} \\[0.5em]
        \bf{0} & (\mC'_{\geq j,\geq j})^{-1}
    \end{pmatrix}.
\end{align}
Thus, the block form of the relation $\Hp = (\Cp^T \Cp)^{-1} = (\Cp)^{-1} (\Cp^T)^{-1} \equiv (\Cp)^{-1} (\Cp)^{-T}$ is
\begin{align}\label{eq:cholesky-block}
    \begin{pmatrix}
        \mH'_{<j,<j} & \mH'_{<j,\geq j} \\
        \mH'_{\geq j,<j} & \mH'_{\geq j,\geq j}
    \end{pmatrix}
    = \begin{pmatrix}
        \star & \star \\[0.5em]
        \bf{0} & (\mC'_{\geq j,\geq j})^{-1}
    \end{pmatrix}
    \begin{pmatrix}
        \star & \bf{0} \\
        \star & (\mC'_{\geq j,\geq j})^{-T}
    \end{pmatrix}
\end{align}
where ``$\star$'' denotes terms that are irrelevant for our proof.
We can read off $\mH'_{\geq j,\geq j}$ from \autoref{eq:cholesky-block},
\begin{align}\label{eq:inverse-h}
    &\mH'_{\geq j,\geq j} = (\mC'_{\geq j,\geq j})^{-1} \,(\mC'_{\geq j,\geq j})^{-T}
    \qquad\Longrightarrow\qquad
    (\mH'_{\geq j,\geq j})^{-1} = (\mC'_{\geq j,\geq j})^T \, \mC'_{\geq j,\geq j}.
\end{align}
Thus, by explicitly evaluating the $(j,k)$-th entry of $(\mH'_{\geq j,\geq j})^{-1}$, we find for all $k \geq j$,
\begin{align}\label{eq:proposition-iii-final}
    \Hipi{j}{jk}
    = \sum_{l=j}^m \Cpi{lj} \Cpi{lk} 
    = \Cpi{jj} \Cpi{jk} 
\end{align}
where only the term with $l=j$ contributes to the sum because $C'_{lj} = 0$ for $l > j$ since $\Cp$ is upper triangular.
Both parts of \autoref{eq:proposition-iii-reduced-claim} follow immediately from \autoref{eq:proposition-iii-final}.

\end{document}